\pdfoutput=1

\documentclass[11pt]{article}
\usepackage[table,xcdraw,dvipsnames]{xcolor}
\usepackage{naacl2021}

\usepackage{times}
\usepackage{latexsym}
\usepackage{relsize}

\usepackage[T1]{fontenc}

\usepackage[utf8]{inputenc}

\usepackage{microtype}
\usepackage{mathtools,amsmath,amsfonts,amssymb}
\usepackage{multirow, tabularx, booktabs, arydshln} 
\usepackage{xspace}
\usepackage{subcaption}
\usepackage{enumitem}

\usepackage{scalerel}
\def\shef{\scalerel*{\includegraphics{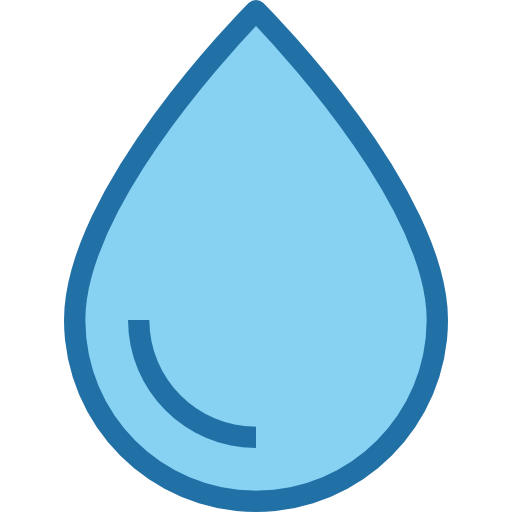}}{\textrm{\textbigcircle}}}
\def\uu{\scalerel*{\includegraphics{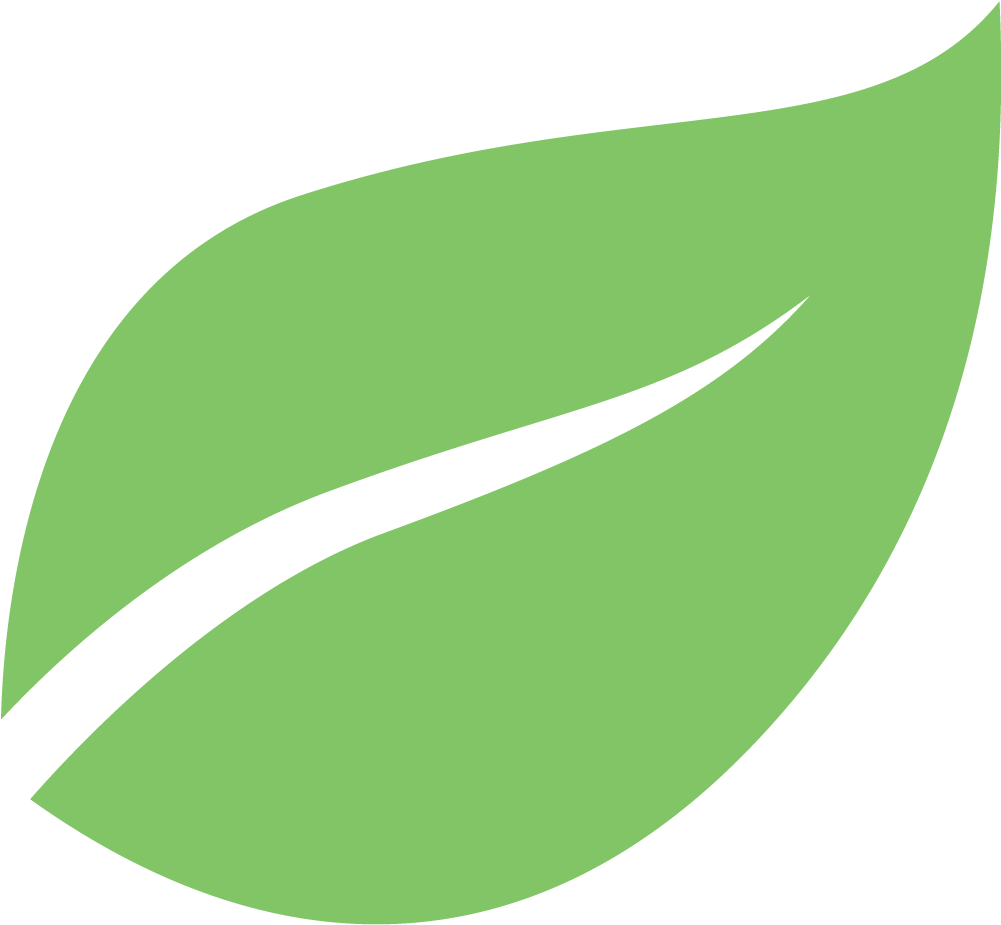}}{\textrm{\textbigcircle}}}
\def\carboon{\scalerel*{\includegraphics{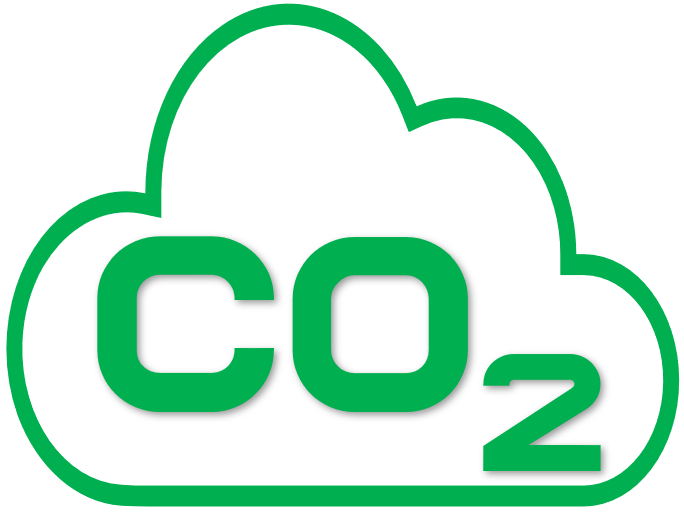}}{\textrm{\textbigcircle}}}
\def\time{\scalerel*{\includegraphics{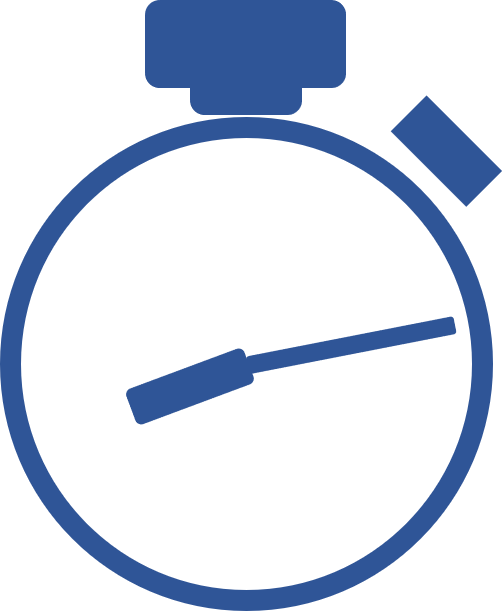}}{\textrm{\textbigcircle}}}

\usepackage{cleveref}
\crefname{section}{§}{§§}
\newcommand{\PE}{\textsc{ProcrustEs}\xspace}
\newcommand{\COtwo}{$\mathrm{CO_2}$\xspace}
\newcolumntype{P}[1]{>{\centering\arraybackslash}p{#1}}
\newcolumntype{Y}{>{\centering\arraybackslash}X}

\title{Highly Efficient Knowledge Graph Embedding Learning \\
with {O}rthogonal {P}rocrustes {A}nalysis}

\author{Xutan Peng\textsuperscript{\shef} \hspace{4mm}  Guanyi Chen\textsuperscript{\uu} \hspace{4mm}  Chenghua Lin\textsuperscript{\shef}\thanks{~~Chenghua Lin is the corresponding author.}~ \hspace{4mm}  Mark Stevenson\textsuperscript{\shef}\\
  \textsuperscript{\shef}Department of Computer Science, The University of Sheffield \\
  \textsuperscript{\uu}Department of Information and Computing Sciences, Utrecht University \\ 
  \texttt{\{x.peng, c.lin, mark.stevenson\}@shef.ac.uk} \hspace{4mm} \texttt{g.chen@uu.nl}
}

\begin{document}
\maketitle
\begin{abstract}
Knowledge Graph Embeddings (KGEs) have been intensively explored in recent years due to their promise for a wide range of applications. 
However, existing studies focus on improving the final model performance without acknowledging the computational cost of the proposed approaches, in terms of execution time and environmental impact. This paper proposes a simple yet effective KGE framework which can reduce the training time and carbon footprint by orders of magnitudes compared with state-of-the-art approaches, while producing competitive performance. We highlight three technical innovations: full batch learning via relational matrices, closed-form Orthogonal Procrustes Analysis for KGEs, and non-negative-sampling training. In addition, as the first KGE method whose entity embeddings also store full relation information, our trained models encode rich semantics and are highly interpretable. Comprehensive experiments and ablation studies involving 13 strong baselines and two standard datasets verify the effectiveness and efficiency of our algorithm.
\end{abstract}

\section{Introduction}

\begin{figure*}[t]
  \centering
  \includegraphics[width=\textwidth, trim={3.5cm .8cm .4cm 3.8cm}, clip]{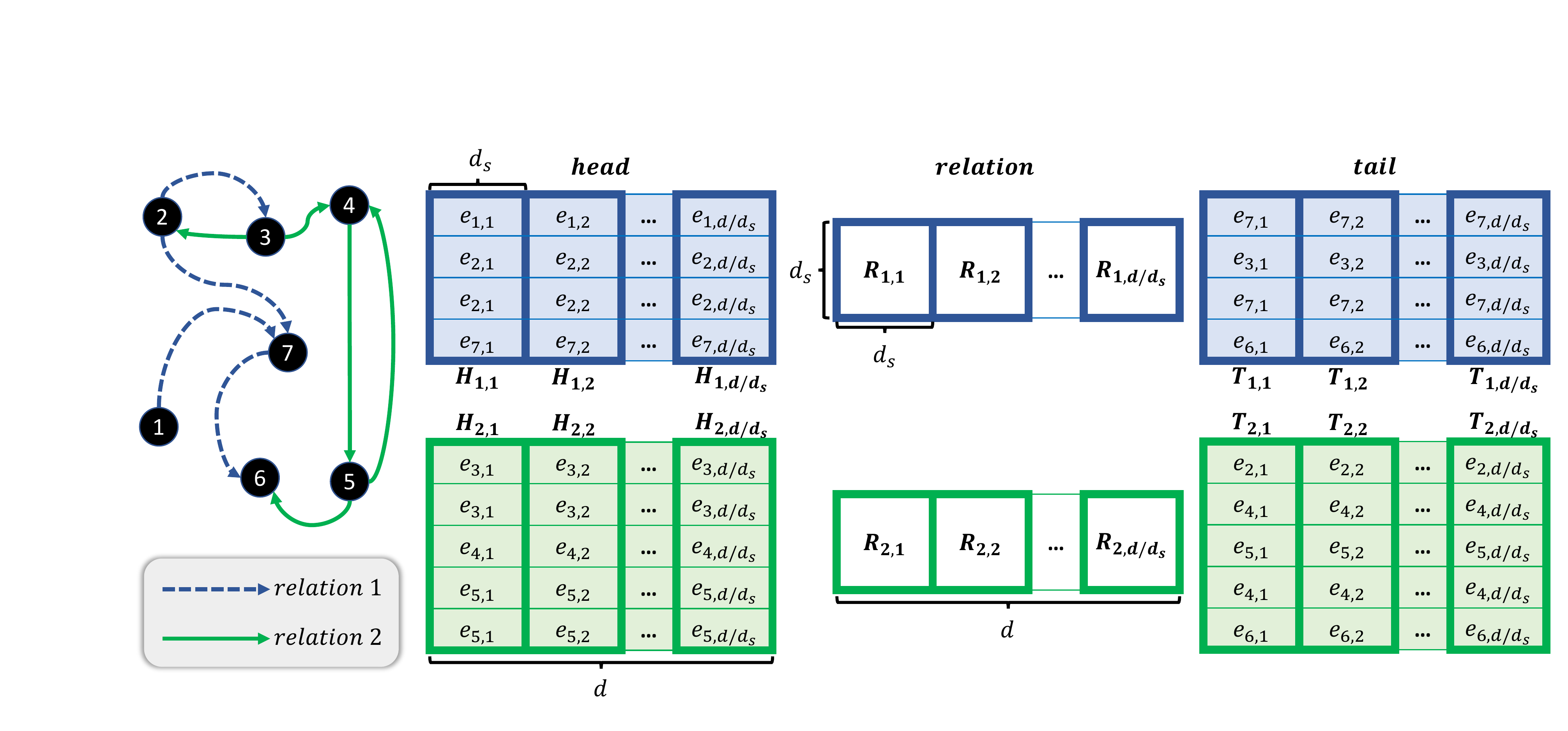} 
\caption{The by-relation partitioning architecture of \PE for a toy graph (left). Matrices involved in the computation of Eq.~\eqref{eq:loss} are divided into two relational matrices: the \textbf{\textcolor{Blue}{upper}} is for \textit{relation} 1 (\textbf{\textcolor{Blue}{dashed}}) and the \textbf{\textcolor{ForestGreen}{lower}} is for \textit{relation} 2 (\textbf{\textcolor{ForestGreen}{solid}}).}
\label{fig:structure}
\end{figure*}

The recent growth in energy requirements for Natural Language Processing (NLP) algorithms has led to the recognition of the importance of computationally cheap and eco-friendly approaches \citep{strubell-etal-2019-energy}. The increase in computational requirements can, to a large extent, be attributed to the popularity of massive pre-trained models, such as Language Models (e.g., BERT~\citep{bert} and GPT-3~\citep{gpt3}) and Knowledge Graph Embeddings (KGEs, e.g., SACN~\citep{SACN}), that require significant resources to train. A number of solutions have been proposed such as reducing the number of parameters the model contains. For instance, \citet{sanh2019distilbert} introduced a distilled version of BERT and \citet{QuatE} decreased the parameters used for training KGEs with the help of the quaternion. In contrast with previous work, this paper explores algorithmic approaches to the development of efficient KGE techniques.

Knowledge Graphs are core to many NLP tasks and downstream applications, such as question answering~\citep{kg-qa}, dialogue agents~\citep{kg-diag}, search engines~\citep{search} and recommendation systems~\citep{recsys}. Facts stored in a knowledge graph are always in the format of tuples consisting of one head entity, one tail entity (both are nodes in knowledge graphs) and a relation (an edge in knowledge graphs) between them. KGEs learn representations of relations and entities in a knowledge graph, which are then utilised in downstream tasks like predicting missing relations~\citep{TransE,RotatE,ote}. The application of deep learning has led to significant advances in KGE~\citep{rossi2020knowledge}. Nonetheless,  such approaches are computationally expensive with associated environmental costs. For example, training the SACN model~\citep{SACN} can lead to emissions of more than 5.3kg \COtwo (for more data of other algorithms, see Tab.~\ref{tab:main}).

To alleviate the computational cost we introduce \PE, a lightweight, fast, and eco-friendly KGE training technique. \PE is built upon three novel techniques. First, to reduce the batch-wise computational overhead, we propose to parallelise batches by grouping tuples according to their relations,  which ultimately enables efficient full batch learning. 
Second, we turn to a closed-form solution for Orthogonal Procrustes Problem to boost the embedding training, which has never been explored in the context of KGEs.
Third, to break though the bandwidth bottleneck, our algorithm is allowed to be trained without negative samples.

To verify the effectiveness and efficiency of our proposed method, we benchmark two popular datasets (WN18RR and FB15k-237) against 13 strong baselines.
Experimental results show that \PE yields performance competitive with the state-of-the-art while also reducing training time by up to 98.4\% and the carbon footprint by up to 99.3\%.
In addition, we found that our algorithm can produce easily interpretable entity embeddings with richer semantics than previous approaches.  
Our code is available at \url{https://github.com/Pzoom522/ProcrustEs-KGE}.

Our contribution is three-fold: 
(1) We introduce three novel approaches to substantially reduce computational overhead of embedding large and complex knowledge graphs: full batch learning based on relational matrices, closed-form Orthogonal Procrustes Analysis for KGEs, and non-negative-sampling training. 
(2) We systemically benchmark the proposed algorithm against 13 strong baselines on two standard datasets, demonstrating that it retains highly competitive performance with just order-of-minute training time and emissions of less than making two cups of coffee.
(3) We successfully encode both entity and relation information in a single vector space for the first time, thereby enriching the expressiveness of entity embeddings and producing new insights into interpretability. 

\section{Methodology}\label{sec:method}

We propose a highly efficient and lightweight method for training KGEs called \PE, which is more efficient in terms of time consumption and \COtwo emissions than previous counterparts by orders of magnitude while retaining strong performance. 
This is achieved by introducing three novel optimisation strategies, namely, relational mini-batch, closed-form Orthogonal Procrustes Analysis, and non-negative sampling training.

\subsection{Preliminaries: Segmented Embeddings}\label{ssec:segment}

Our proposed \PE model is built upon \textit{segmented embeddings}, a technique which has been leveraged by a number of promising recent approaches to KGE learning (e.g., RotatE~\citep{RotatE}, SEEK~\citep{SEEK}, and OTE~\citep{ote}). 
In contrast to conventional methods for KGEs where each entity only corresponds to one single vector, algorithms adopting segmented embeddings explicitly divide the entity representation space into multiple independent sub-spaces. During training each entity is encoded as a concatenation of decoupled sub-vectors (i.e., different segments, and hence the name). For example, as shown in Fig.~\ref{fig:structure}, to encode a graph with 7 entities,  the embedding of the $t$th entity is the row-wise concatenation of its $d/d_s$ sub-vectors (i.e., ${e_{t,1}}^\frown{e_{t,2}}^\frown \ldots ^\frown{e_{t,d/d_s}}$), where $d$ and $d_s$ denote the dimensions of entity vectors and sub-vectors, respectively. 
Employing segmented embeddings permits parallel processing of the structurally separated sub-spaces, and hence significantly boosts the overall training speed. Furthermore, segmented embeddings can also enhance the overall expressiveness of our model, while substantially reducing the dimension of matrix calculations. We provide detailed discussion on the empirical influence of segmented embedding setups in \cref{subsec:exp_dim}.

\subsection{Efficient KGE Optimisation}

\paragraph{Full batch learning via relational matrices.}
Segmented embeddings can speed up training process by parallelising tuple-wise computation. In this section, we propose a full batch learning technique via relational matrices, which can optimise batch-wise computation to further reduce training time.  
This idea is motivated by the observation that existing neural KGE frameworks all perform training based on random batches constructed from tuples consisting of different types of relations~\cite{TransE,ComplEx,R-GCN,low-acl20}.
Such a training paradigm is based on random batches which, although straightforward to implement, is difficult to parallelise.
This is due to the nature of computer process scheduling: during the interval between a process reading and updating the relation embeddings, they are likely to be modified by other processes, leading to synchronisation errors and consequently result in unintended data corruption, degraded optimisation, or even convergence issues.

To tackle this challenge, we propose to construct batches by grouping tuples which contain \textit{the same relations}.
The advantage of this novel strategy is two-fold. For one thing, it naturally reduces the original tuple-level computation to simple matrix-level arithmetic. For another and more importantly, we can then easily ensure that the embedding of each relation is only accessible by \textit{one single process}. Such a training strategy completely avoids the data corruption issue. In addition, it makes the employment of the full batch learning technique (via relational matrices) possible, which offers a robust solution for parallelising the KGEs training process and hence can greatly enhance the training speed. 
To the best of our knowledge, this approach has never been explored by the KGE community.

As illustrated in Fig.~\ref{fig:structure}, we first separate the embedding space into segments (cf. \cref{ssec:segment}) and arrange batches based on relations. After that, for each training step, the workflow of \PE is essentially decomposed into $m \times d/d_s$ parallel optimisation processes, where $m$ is the number of relation types.
Let $i$ and $j$ denote the indices of relation types and sub-spaces, respectively, then the column-wise concatenations of the $j$th sub-vectors of all tuples of $i$th relations can be symbolised as $H_{i,j}$ (for head entities) and $T_{i,j}$ (for tail entities). Similarly, $R_{i,j}$ denotes the corresponding relation embedding matrix in the $j$th sub-space. The final objective function of \PE becomes
\begin{equation}\label{eq:loss}
   \mathcal{L} =  \sum_{i=1}^{m} \sum_{j=1}^{d/d_s} || H_{i,j} R_{i,j} - T_{i,j}||_2. 
\end{equation}

\paragraph{Orthogonal Procrustes Analysis.}

Our key optimisation objective, as formulated in Eq.~\eqref{eq:loss}, is to minimise the Euclidean distance between the head and tail matrices for each parallel process. 
In addition, following \citet{RotatE} and \citet{ote}, we restrict the relation embedding matrix $R_{i,j}$ to be orthogonal throughout model training, which has been shown effective in improving KGE quality.
Previous KGE models use different approaches to impose orthogonality.
For instance, RotatE~\citep{RotatE} takes advantage of a corollary of Euler’s identity and defines its relation embedding as
\begin{equation}\label{eq:rotate}
     R_{i,j} = \begin{bmatrix}
    \cos{\theta_{i,j}} &\! \sin{\theta_{i,j}} \\
    -\sin{\theta_{i,j}} &\! \cos{\theta_{i,j}}
    \end{bmatrix},
\end{equation}
which is controlled by a learnable parameter $\theta_{i,j}$. Although Eq.~\eqref{eq:rotate} holds orthogonality and retains simplicity, it is essentially a special case of segmented embedding where $d_s$ equals 2. As a result, $R_{i,j}$ is always two-dimensional, which greatly limits the modelling capacity (see \cref{subsec:exp_dim} for discussion on the impact of dimensionality).
To overcome this limitation, OTE~\citep{ote} explicitly orthogonalises $R_{i,j}$ using the Gram-Schmidt algorithm per back-propagation step (see Appendix~\ref{app:Gram-Schmidt} for details). However, while this scheme works well for a wide range of $d_s$ (i.e., the dimension for the sub-vector), similar to RotatE, OTE finds a good model solution based on gradient descent, which is computationally very expensive.

We address the computational issue by proposing a highly efficient method  utilising the proposed parallelism of full batch learning. With full batch learning, comparing with existing methods which deal with heterogeneous relations, \PE only needs to optimise one single $R_{i,j}$ in each process, which becomes a simple constrained matrix regression task. More importantly, through Singular Value Decomposition (SVD), we can derive an \textit{closed-form} solution~\citep{opp}  as
\begin{equation}\label{eq:opp_solution} 
    R_{i,j}^{\star} = U V^\intercal, \text{w/ } U\Sigma V^\intercal = \mathrm{SVD}(H_{i,j}^\intercal T_{i,j}),
\end{equation}
where $R_{i,j}^{\star}$ denotes the optima. During each iteration, \PE can directly find the \textit{globally} optimal embedding for each relation given the current entity embeddings by applying Eq.~\eqref{eq:opp_solution}. Then, based on the calculated $\mathcal{L}$, \PE updates entity embeddings through the back propagation mechanism (NB: the relation embeddings do not require gradients here). This process is repeated until convergence. As the optimisation of relation embeddings can be done almost instantly per iteration thanks to the closed-form Eq.~\eqref{eq:opp_solution}, \PE is significantly (orders of magnitude) faster than RotatE and OTE. In addition, compared with entity embeddings of all other KGE models which are updated separately with relation embedding, entity embeddings trained by \PE can be used to restore relation embeddings directly (via Eq.~\eqref{eq:opp_solution}). In other words, \PE can encode richer information in the entity space than its counterparts (see \cref{subsec:interpretability}).

\paragraph{Further optimisation schemes.}
As recently surveyed by \citet{dog}, existing KGE methods employ negative sampling as a standard technique for reducing training time, 
where update is performed only on a subset of parameters by calculating loss based on the generated negative samples. 
With our proposed closed-form solution (i.e., Eq.~\eqref{eq:opp_solution}), computing gradients to update embeddings is no longer an efficiency bottleneck for \PE.
Instead, the speed bottleneck turns out to be the extra bandwidth being occupied due to the added negative samples.
Therefore, for \PE, we do not employ negative sampling but rather update all embeddings during each round of back propagation with positive samples only, in order to further optimise the training speed (see Appendix~\ref{app:bandwidth} for bandwidth comparisons against baselines which adopts negative sampling). 

We also discovered that if we do not apply any additional conditions during training, \PE tends to fall into a trivial optimum after several updates, i.e., $\mathcal{L}=0$, with all values in $H_{i,j}$, $T_{i,j}$ and $R_{i,j}$ being zero. In other words, the model collapses with nothing encoded at all. This is somewhat unsurprising as such trivial optima often yields large gradient and leads to this behaviour~\citep{trivial}. 
To mitigate this degeneration issue, inspired by the geometric meaning of orthogonal $R_{i,j}$ (i.e., to rotate $H_{i,j}$ towards $T_{i,j}$ around the coordinate origin, without changing vector length), we propose to constrain all entities to a high-dimensional \textit{hypersphere} by performing two spherisation steps in every epoch. 
The first technique, namely \textit{centring}, respectively translates $H_{i,j}$ and $T_{i,j}$ so that the column-wise sum of each matrix becomes a zero vector (note that each row denotes a sub-vector of an entity).
The second operation is \textit{length normalisation}, which ensures the row-wise Euclidean norm of $H_{i,j}$ and $T_{i,j}$ to always be one. Employing these two simple constraints effectively alleviates the trivial optimum issue, as evidenced in our experiments (see \cref{sec:exp}).  

\section{Experiment}\label{sec:exp}

\subsection{Setups}

We assess the performance of \PE on the task of multi-relational link prediction, which is the \textit{de facto} standard of KGE evaluation.

\begin{table}[t]
    \centering \small
    \begin{tabular}{lp{3cm}<{\centering}c}
    \toprule
          &  FB15k-237 & WN18RR \\ 
    \midrule
         Entities & 14,541 & 40,943 \\
         Relations & 237 & 11 \\
         Train samples & 272,115 & 86,835  \\
         Validate samples & 17,535 & 3,034 \\
         Test samples & 20,466 & 3,134 \\
         \bottomrule
    \end{tabular}
    \caption{Basic statistics of the two benchmark datasets.}
    \label{tab:dataset}
\end{table}
\newcommand{\rlkt}{\textcolor{BlueGreen!15}{*}}
\newcommand{\wlkt}{\textcolor{white}{*}}
\begin{table*}[t]\small
\centering
\begin{tabular}{{p{2.7cm}:P{1.00cm}P{1.00cm}P{1.00cm}P{1.00cm}:P{1.00cm}:P{1.00cm}:P{1.00cm}P{1.00cm}P{1.00cm}P{1.00cm}:P{1.00cm}:P{1.00cm}}}
\toprule 
\multirow{2}{*}{} & \multicolumn{6}{c:}{\textbf{WN18RR}} & \multicolumn{6}{c}{\textbf{FB15k-237}} \\ \cline{2-13} 
  & MRR & H1 & H3 & H10 & \time & \carboon & MRR & H1 & H3 & H10  & \time & \carboon  \\ \midrule 
TransE (\citeyear{TransE}) & \cellcolor{BlueGreen!40} .226 & - & - & \cellcolor{BlueGreen!40}.501  & 85 & 367 & \cellcolor{BlueGreen!40}.294 & - & - & \cellcolor{BlueGreen!15}.465 & 96 & 370 \\ 
DistMult (\citeyear{DistMult}) & \cellcolor{BlueGreen!40}.430 & \cellcolor{BlueGreen!40}.390 & \cellcolor{BlueGreen!40}.440 & \cellcolor{BlueGreen!40}.490 & 79 & 309 & \cellcolor{BlueGreen!40}.241 & \cellcolor{BlueGreen!40}.155 & \cellcolor{BlueGreen!40}.263 & \cellcolor{BlueGreen!40}.419 & 91 & 350 \\ 
ComplEx (\citeyear{ComplEx}) & \cellcolor{BlueGreen!40}.440 & \cellcolor{BlueGreen!15}.410 & \cellcolor{BlueGreen!40}.460 & \cellcolor{BlueGreen!40}.510 & 130 & 493 &  \cellcolor{BlueGreen!40}.247 & \cellcolor{BlueGreen!40}.158 & \cellcolor{BlueGreen!40}.275 & \cellcolor{BlueGreen!40}.428 & 121 & 534 \\ 
R-GCN (\citeyear{R-GCN}) & \cellcolor{BlueGreen!40}.417 & \cellcolor{BlueGreen!40}.387 & \cellcolor{BlueGreen!40}.442 & \cellcolor{BlueGreen!40}.476 & 138 & 572 & \cellcolor{BlueGreen!40}.248 & \cellcolor{BlueGreen!40}.151 & \cellcolor{BlueGreen!40}.264 & \cellcolor{BlueGreen!40}.417 & 152 & 598 \\ 
ConvE (\citeyear{ConvE}) & \cellcolor{BlueGreen!40}.430 & \cellcolor{BlueGreen!40}.400 & \cellcolor{BlueGreen!40}.440 & \cellcolor{BlueGreen!40}.520 & 840 & 3702 & \cellcolor{BlueGreen!15}.325 & \cellcolor{BlueGreen!40}.237 & \cellcolor{BlueGreen!15}.356 & \cellcolor{BlueGreen!15}.501 & 1007 & 4053 \\ 
A2N (\citeyear{A2N}) & \cellcolor{BlueGreen!40}.450 & \cellcolor{BlueGreen!15}.420 & \cellcolor{BlueGreen!40}.460 & \cellcolor{BlueGreen!40}.510 & 203 & 758 & \cellcolor{BlueGreen!15}.317 & \cellcolor{BlueGreen!40}.232 & \cellcolor{BlueGreen!15}.348 & \cellcolor{BlueGreen!15}.486 & 229 & 751 \\
SACN (\citeyear{SACN}) & \cellcolor{BlueGreen!15}.470 & .430 & \cellcolor{BlueGreen!40}.480 & \cellcolor{BlueGreen!40}.540 & 1539 & 5342 & .352 & .261 & .385 & \cellcolor{BlueGreen!15}.536 & 1128 & 4589 \\ 
TuckER (\citeyear{TuckER}) & \cellcolor{BlueGreen!15}.470 & .443 & \cellcolor{BlueGreen!40}.482 & \cellcolor{BlueGreen!40}.526 & 173 & 686 & \textbf{.358} & \textbf{.266} & \textbf{.392} & .544 & 184 & 704 \\ 
QuatE (\citeyear{QuatE}) & .488 & .438 & .508 & .582 & 176 & 880 & .348 & \cellcolor{BlueGreen!15}.248 & .382 & \textbf{.550} & 180 & 945 \\
InteractE (\citeyear{InteractE}) & \cellcolor{BlueGreen!15}.463 & .430 & - & \cellcolor{BlueGreen!40}.528 & 254 & 1152 & .354 & .263 & - & \cellcolor{BlueGreen!15}.535 & 267 & 1173 \\
RotH (\citeyear{low-acl20}) & \textbf{.496} & \textbf{.449} & \textbf{.514} & \textbf{.586} & 192 & 903 & \cellcolor{BlueGreen!15}.344 & \cellcolor{BlueGreen!15}.246 & .380 & \cellcolor{BlueGreen!15}.535 & 207 & 1120 \\ \hdashline 
RotatE (\citeyear{RotatE}) & \cellcolor{BlueGreen!40}.439 & \cellcolor{BlueGreen!40}.390 & \cellcolor{BlueGreen!40}.456 & \cellcolor{BlueGreen!40}.527 & 255 & 823 & \cellcolor{BlueGreen!15}.297 & \cellcolor{BlueGreen!40}.205 & \cellcolor{BlueGreen!15}.328 & \cellcolor{BlueGreen!15}.480 & 343 & 1006 \\
OTE (\citeyear{ote}) & \cellcolor{BlueGreen!40} .448 & \cellcolor{BlueGreen!40}.402 & \cellcolor{BlueGreen!40}.465 & \cellcolor{BlueGreen!40}.531 & 304 & 1008 & \cellcolor{BlueGreen!15}.309 & \cellcolor{BlueGreen!40}.213 & \cellcolor{BlueGreen!15}.337 & \cellcolor{BlueGreen!15}.483  & 320 & 1144 \\
\midrule \midrule
\cellcolor{BlueGreen!40}\PE (ours) & .453 & .408 & .491 & .549 & \textcolor{Blue}{\textbf{14}} & \textcolor{ForestGreen}{\textbf{37}} & .295 & .241 & .310 & .433 & \textcolor{Blue}{\textbf{9}} & \textcolor{ForestGreen}{\textbf{42}} \\\hdashline
w/ \textsc{NS} (ours) & .457 & .411 & .494 & .551  & 44 & 124 & .302 & .245 & .333 & .465  & 37 & 159 \\
w/ \textsc{TB} (ours) & .468 & .417 & .498 & .557  & 92 & 268 & .326 & .247 & .354 & .492  & 56 & 243 \\
\cellcolor{BlueGreen!15} w/ \textsc{NS+TB} (ours) & .474 & .421 & .502 & .569  & 131 & 346 & .345 & .249 & .379 & .541  & 85 & 285 \\ 
\bottomrule
\end{tabular}
\caption{Model effectiveness and efficiency on link prediction benchmarks. \time: training time (\textit{minutes}); \carboon: carbon dioxide production (\textit{grams}). \textsc{NS}: negative sampling; \textsc{TB}: traditional batch. The performance results of baselines are coloured \colorbox{BlueGreen!40}{heavily} and \colorbox{BlueGreen!15}{lightly} if they are below those of \PE and ``w/ \textsc{NS+TB}'', respectively.
State-of-the-art scores are in \textbf{bold}. 
Following \citet{TuckER} and \citet{QuatE}, for fair comparison, both RotatE and OTE results are reported with conventional negative sampling rather than the self-adversarial one. 
}
\label{tab:main}
\end{table*}

\paragraph{Datasets.}~~In this study, following previous works (e.g., baselines in Tab.~\ref{tab:main}), we employ two benchmark datasets for link prediction: (1) FB15K-237~\citep{toutanova-chen-2015-observed}, which consists of sub-graphs extracted from Freebase, and contains no inverse relations; and (2) WN18RR~\citep{ConvE}, which is extracted from WordNet. Tab.~\ref{tab:dataset} shows descriptive statistics for these two datasets, indicating that FB15K-237 is larger in size and has more types relations while WN18RR has more entities. We use the same training, validating, and testing splits as past studies. 

\paragraph{Evaluation metrics.}~~Consistent with~\citet{RotatE} and~\citet{ote}, we report Hit Ratio with cut-off values $n=1, 3, 10$ (i.e., H1, H3, and H10) and Mean Reciprocal Rank (MRR). 
Additionally, as to efficiency, we report the time cost and \COtwo emissions for each model, i.e., from the beginning of training until convergence. 

\paragraph{Baselines.}~~We compare \PE to not only classical neural graph embedding methods, including TransE~\citep{TransE}, DistMulti~\citep{DistMult}, and ComplEx~\citep{ComplEx}, but also embedding techniques recently reporting state-of-the-art performance on either WN18RR or FB15k-237, including R-GCN~\citep{R-GCN}, ConvE~\citep{ConvE}, A2N~\citep{A2N}, RotatE~\citep{RotatE}, SACN~\citep{SACN}, TuckER~\citep{TuckER}, QuatE~\citep{QuatE}, InteractE~\citep{InteractE}, OTE~\citep{ote}, and RotH~\citep{low-acl20}. 
For all these baselines, we use the official code and published hyper-parameters to facilitate reproducibility.

\paragraph{Implementation details.}
All experiments are conducted on a workstation with one NVIDIA GTX 1080 Ti GPU and one Intel Core i9-9900K CPU, which is widely applicable to moderate industrial/academic environments. We use the Experiment Impact Tracker~\citep{tracker} to benchmark the time and carbon footprint of training. To reduce measurement error, in each setup we fix the random seeds, run \PE and all baselines for three times and reported the average.

The key hyper-parameters of our model is $d$ and $d_s$, which are respectively set at 2K and 20 for both datasets. The detailed selection process is described in \cref{subsec:exp_dim}.
We train each model for a maximum of 2K epochs and check if the validation MRR stops increasing every 100 epochs after 100 epochs. For WN18RR and FB15k-237 respectively, we report the best hyperparameters as fixed learning rates of 0.001 and 0.05 (Adam optimiser), and stopping epochs of 1K and 200.

\subsection{Main Results}\label{ssec:main_results}

\begin{figure*}[t]
  \centering
  \includegraphics[width=\textwidth, trim={.4cm .0cm .4cm .0cm}, clip]{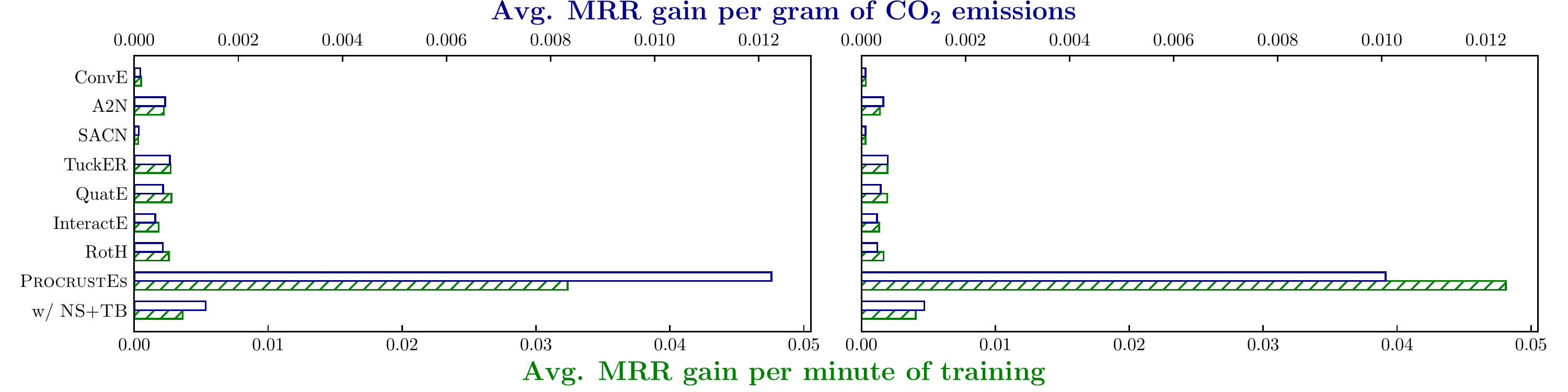}  
\caption{Unified effectiveness-efficiency comparison between most competitive KGE models in Tab.~\ref{tab:main}. The left and right sub-figures are respectively for WN18RR and FB15k-237.}
\label{fig:unit}
\end{figure*}

Tab.~\ref{tab:main} reports the results of both our \PE and all other 13 baselines on both WN18RR and FB15k-237 datasets. We analyse these results 
from two dimensions: 
(1) \textbf{Effectiveness}: the model performance on link prediction task (MRR is our main indicator);
(2) \textbf{Efficiency}: system training time and carbon footprint (i.e., \COtwo emissions).

Regarding the performance on WN18RR, we found that \PE performs as good as or even better than previous state-of-the-art approaches.
To be concrete, out of all 13 baselines, it beats 11 in H10, (at least) 9 in H3 and 8 in MRR. The models outperformed by \PE include not only all methods prior to 2019, but also several approaches published in 2019 or even 2020. 
Notably, when compared with the RotatE and OTE, two highly competitive methods which have similar architectures to \PE (i.e., with segmented embeddings and orthogonal constraints), our \PE  can learn KGEs with higher quality (i.e., 0.014 and 0.005 higher in MRR, respectively). This evidences the effectiveness of the proposed approaches in \cref{sec:method} in modelling knowledge tuples. 

While \PE achieves very competitive performance, it requires significantly less time for training: it converges in merely \textbf{14 minutes}, more than 100 times faster than strong-performing counterparts such as SACN. Moreover, it is very environmentally friendly: from bootstrapping to convergence, \PE only emits \textbf{37g} of \COtwo, which is even less than making two cups of coffee\footnote{\url{https://tinyurl.com/coffee-co2}}. On the contrary, the baselines emit on average 1469g and up to 5342g \COtwo: the latter is even roughly equal to the carbon footprint of a coach ride from Los Angeles to San Diego\footnote{\url{https://tinyurl.com/GHG-report-2019}}.

As for the testing results on FB15k-237, we found that although \PE seems less outstanding (we investigate the reasons in \cref{ssec:ablation}), it still outperforms at least 7 more complex baselines in H1 and almost all models prior to 2019 in MRR. Furthermore, similar to the observation on WN18RR, it demonstrates great advantage in terms of efficiency. While all baselines need 91 to 1128 minutes to coverage with 350g to 4589g \COtwo produced, \PE can learn embeddings of similar quality in just \textbf{9 minutes} and with \textbf{42g emissions}. 
By employing both traditional batch and negative sampling, we show that \PE can achieve near-state-of-the-art performance on both datasets. We discuss this in detail in \cref{ssec:ablation}.

To provide a unified comparisons between \PE and the most strong-performing baselines on both effectiveness and efficiency, we further investigate the following question: How much performance gain can we obtain by spending \textit{unit} time on training or making \textit{unit} emissions? We did analysis by calculating MRR/(training time) and MRR/(carbon footprint) and the results are presented in Fig.~\ref{fig:unit}.
It is obvious that among all competitive KGE models, \PE is the most economic algorithm in terms of performance-cost trade-off: it is \textit{more than 20 times} more efficient than any past works, in terms of both performance per unit training time and per unit \COtwo emissions. 

We also investigate baseline performance with a shorter training schedule. From scratch, we train RotH, the best performing algorithm on WN18RR, and stop the experiment when MRR reaches the performance of \PE. On WN18RR, RotH takes 50 minutes (3.6$\times$ \PE) and emits 211g \COtwo (5.7$\times$ \PE); on FB15k-237 RotH takes 45 minutes (5.0$\times$ \PE) and emits 218g \COtwo (5.2$\times$ \PE). These results once again highlight the efficiency superiority of our approach.

\subsection{Ablation Studies}\label{ssec:ablation}

To better understand the performance difference of \PE on WN18RR and FB15k-237, we dive deeply into the dataset statistics in Tab.~\ref{tab:dataset}. 
\citet{full-batch-bad1} and \citet{full-batch-bad2} found that although full batch learning can boost training speed and may benefit performance, when the data distribution is too sparse, it may be trapped into sharp minimum. 
As the average number of samples linked to each relation is significantly smaller for FB15k-237 than for WN18RR (1148 \textit{vs} 7894), the distribution of the former is likely to be more sparse and the generalisability of \PE may thus be harmed.
For another, FB15k-237 has finer-grained relation types (237 \textit{vs}. 11 of WN18RR), so intuitively the likelihood of tuples sharing similar relations rises. However, as \PE omits negative sampling to trade for speed, sometimes it maybe be less discriminative for look-alike tuples.

To validate the above hypotheses, we additionally conduct ablation studies by switching back to traditional batch mode and/or adding negative sampling modules\footnote{Following \citet{RotatE}, we set the batch size at 1024 and the negative sample size at 128.}.
Configurations where the closed-form optimisation, Eq.~\eqref{eq:opp_solution}, is replaced by gradient descent are omitted since the resulting architecture is very similar to OTE.
As shown in the lower section of Tab.~\ref{tab:main}, both using either traditional or negative sampling (i.e., w/ \textsc{NS} and w/ \textsc{TB}) can improve the performance of \PE for all metrics.
For example, on WN18RR our approach (w/ \textsc{NS+TB}) outperforms most baselines and is close to the performance of QuatE and RotH, but thanks to the Orthogonal Procrustes Analysis, the computational cost of our approach is significantly less. Compared to WN18RR, the gain of our model on FB15k-237 by adopting negative sampling and traditional batch is even more significant, achieving near-state-of-the-art performance (i.e., compared to TuckER, the MRR is only 1.3\% less with merely 4.9\% of the computational time). 
These observations verify our aforementioned hypotheses. 
We also found out that traditional batch is more effective than negative sampling for \PE in terms of improving model performance. 
On the other hand, however, adding these two techniques can reduce the original efficiency of \PE to some extend. 

Nevertheless, as Eq.~\eqref{eq:opp_solution} is not only fast but also energy-saving (as only basic matrix arithmetic on GPUs is involved), even \PE with the ``w/ \textsc{NS+TB}'' configuration preserves great advantage in training time and carbon footprint. Moreover, it achieves near-state-of-the-art effectiveness on both datasets (cf. Tab.~\ref{tab:main}) and still exceeds strong baselines in training efficiency with large margins (cf. Fig.~\ref{fig:unit}). One interesting observation is that, while the training time of RotH is merely 1.47$\times$ of that of \PE (w/ \textsc{NS+TB}), their emission levels are drastically different. This is because RotH implements 24-thread multiprocessing by default while our approach creates only one process. Within similar training time, methods like RotH will thus consume a lot more power and emit a lot more \COtwo. Therefore, for effectiveness-intensive applications, we recommend training \PE in transitional batches with negative sampling, as it can then yield cutting-edge performance without losing its eco-friendly fashion.

\subsection{Impacts of Dimensionality}\label{subsec:exp_dim}

\begin{figure}[t]
\begin{subfigure}{\columnwidth}
  \centering
  \includegraphics[width=\columnwidth, trim={.4cm .4cm .4cm .3cm}, clip]{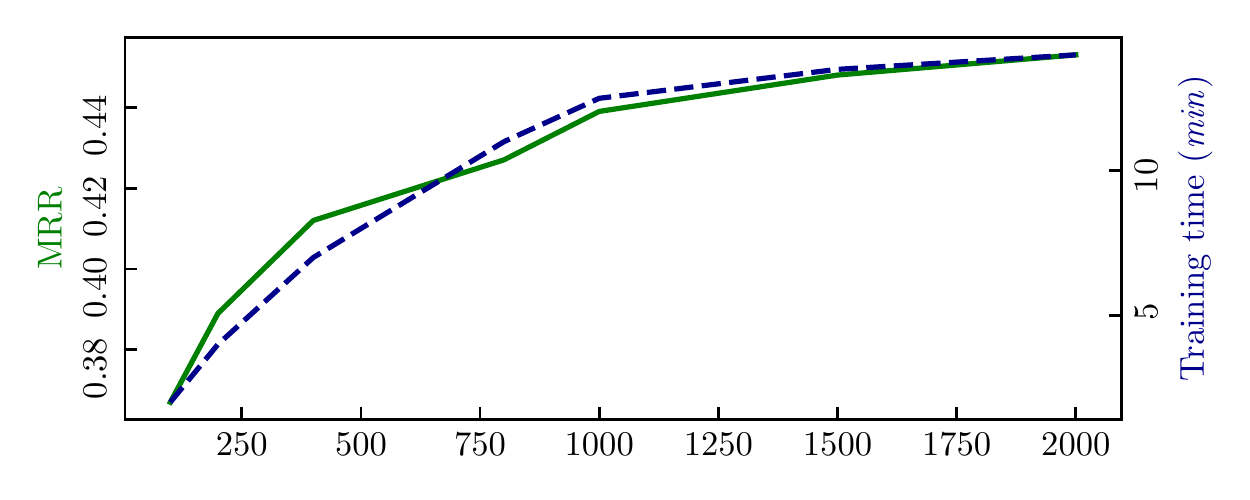}  
\end{subfigure}
\begin{subfigure}{\columnwidth}
  \centering
  \includegraphics[width=\columnwidth, trim={.4cm .4cm .4cm .3cm}, clip]{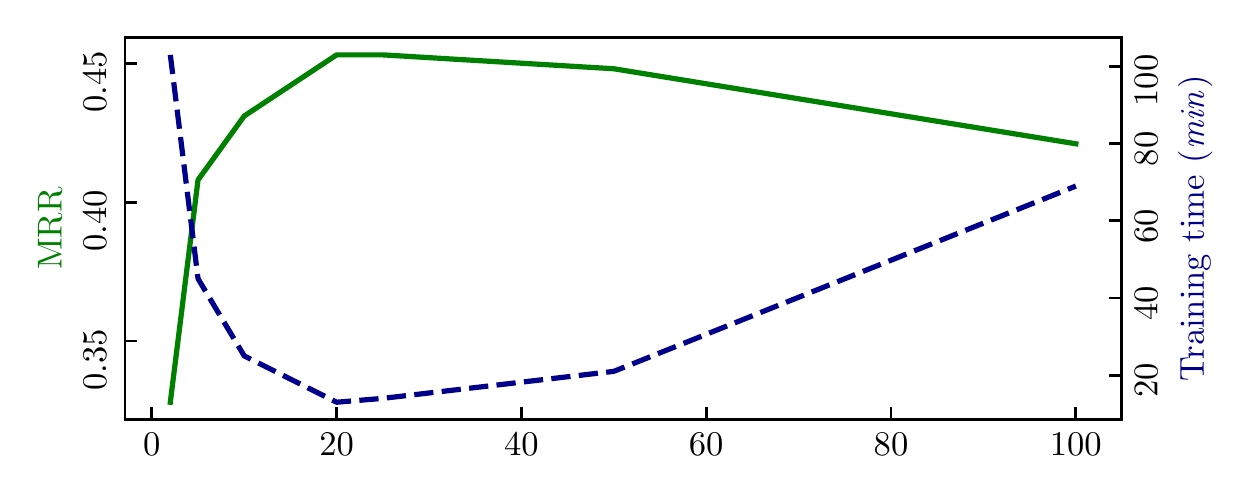}  
\end{subfigure}
\caption{With different $d$ (upper) and $d_s$ (lower), the training time and convergence MRR of \PE on WN18RR (results on FB15k-237 exhibit similar trends). X-axes denote dimensionality.}
\label{fig:dim}
\end{figure}

Our experiments also indicate that the selection of two dimensional hyper-parameters has substantial influence on both effectiveness and efficiency of \PE.
For the dimension of the entire embedding space, we follow the recommendation of \citet{ote} and set $d_s$ at 20. 
We then train \PE with $d \in \{100, 200, 400, 800, 1\mathrm{K}, 1.5\mathrm{K}, 2\mathrm{K}\}$ and plotted results based on the validation set, as shown in Fig.~\ref{fig:dim}. It is evident that with the increase of $d$, the model performance (indicated by MRR) grows but the training time also rises. Observing the curvature of training time almost saturates when $d \geqslant$ 1K, we decide 2K as the best setting for both WN18RR and FB15k-237 given the 11GB graphics memory limit of our hardware.
For the dimension of sub-embeddings, we fix $d$ at 2K and enumerated $d_s \in \{2, 5, 10, 20, 25, 50, 100\}$. For algorithm performance, the pattern we witnessed is on par with that reported by \citet{ote}, i.e., before $d_s$ reaches 20 or 25 the effectiveness jumps rapidly, but after that the model slowly degrades, as the learning capacity of the network reduces. Coincidentally, the training speed also climbs its peak when $d_s$ is 20, making it indisputably become our optimal choice.

\newcolumntype{L}[1]{>{\raggedright\let\newline \\\arraybackslash\hspace{0pt}}m{#1}}
\begin{figure*}[t]
\begin{subfigure}{\columnwidth}
  \centering
  \includegraphics[width=\columnwidth, trim={5.2cm 6.3cm 2.5cm 3.9cm}, clip]{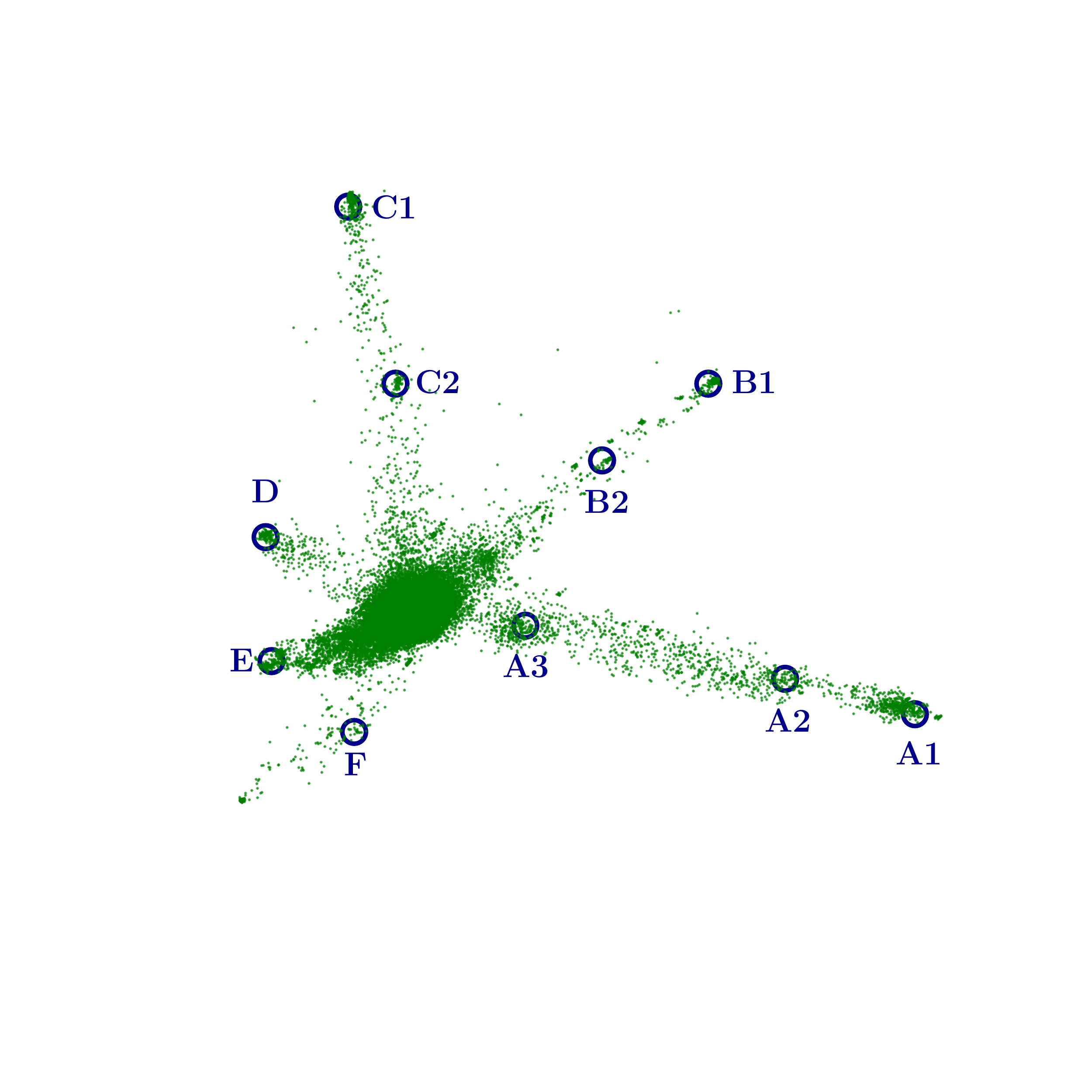} 
\end{subfigure}
\begin{subtable}{\columnwidth}\footnotesize
\begin{tabularx}{8.1cm}{L{0.5cm}X}
 \textbf{A1} & chittagong, cartagena, pittsburgh$\_$of$\_$the$\_$south, le$\_$havre, nanning, stuttgart, kolkata, houston, windy$\_$city, $\ldots$ \\
 \textbf{A2} &  yellowstone$\_$river, atlas$\_$mountains,  san$\_$fernando$\_$valley, sambre$\_$river, nile$\_$river, susquehanna$\_$river, rhine$\_$river, $\ldots$ \\ 
 \textbf{A3} & sudan, balkanshe$\_$alps, east$\_$malaysia, lower$\_$egypt, kalimantan, turkistan, tobago, lowlands$\_$of$\_$scotland, sicily, $\ldots$ \\
  \textbf{B1} &  mefoxin, metharbita, valium,  amobarbital, procaine, nitrostat, tenormin,  minor$\_$tranquillizer, cancer$\_$drug, $\ldots$ \\
 \textbf{B2} & epinephrine, steroid$\_$hormone, internal$\_$secretion, alkaloid, gallamine, prolactin, luteinizing$\_$hormone, $\ldots$ \\
 \textbf{C1} & military$\_$formation, retreat, tactics, strategic$\_$warning, peacekeeping$\_$operation,  unauthorized$\_$absence, $\ldots$ \\
 \textbf{C2} &  commando, sailor$\_$boy, outpost, saddam's$\_$martyrs, military$\_$advisor, battlewagon, commander, $\ldots$ \\
 \textbf{D} &  plaintiff, remitment, franchise, summons, false$\_$pretens, suspect, amnesty, legal$\_$principle, disclaimer, affidavit, $\ldots$ \\
 \textbf{E} &   genus$\_$ambrosia, gloxinia, saintpaulia, genus$\_$cestrum, genus$\_$eriophyllum, valerianella, genus$\_$chrysopsis, $\ldots$ \\
 \textbf{F} &  moneyer, teacher, researcher, president, prime$\_$minister, wheeler$\_$dealer, house$\_$servant, victualler, burglar, $\ldots$ \\
\end{tabularx}
\end{subtable}
\caption{3D PCA visualisation of \PE entity embeddings for WN18RR.}
\label{fig:pca}
\end{figure*}
\subsection{Interpreting Entity Embeddings}\label{subsec:interpretability}

Building on the fact that \PE marry entity information and relation information (in other words, for a specific entity, the information of the entity itself and of its corresponding relations is encoded in a single vector), the location of a entity is more expressive and, thus, the related entity embedding is more interpretable. Picking up on that, we do visualisation study on the trained entity embeddings. To this end, we conduct dimension reduction on the embeddings using Principal Components Analysis (PCA), which reduces the dimensionality of an entity embedding from 2K to three\footnote{We disable axes and grids for visualisation's clarity. Please see the original figure in Appendix~\ref{app:vis_w2v}.}. Fig.~\ref{fig:dim} shows the visualisation result, from which we see a diagram with 6 ``arms''. This is far distinct from the distributional topology of conventional semantic representations, e.g., word embeddings~\citep{w2v} (see Appendix~\ref{app:vis_w2v}).

In Fig.~\ref{fig:dim}, we also list the representative entities that fall in some clusters on each arm. Each cluster is referred by an ID (from A1 to F2). When we zoom into this list, we observe something interesting: \textbf{First}, entities on the same arm are semantically similar, or, in other words, these entities belong to the same category. Concretely, entities on arm A are locations, those on arm B are biochemical terms, and those on arm C are military related entities. Entities on arm D, E, and F consists of entities refer to concepts of law, botany, and occupation, respectively. \textbf{Second}, significant differences exist between each cluster/position on a arm. One example is that, for arm A, A1 are entities for cities, such as \textit{Stuttgart}, \textit{Houston}, \textit{Nanning}; A2 is about entities for rivers, mountains, etc.; and A3 contains entities referring to countries or regions. Similarly, while B1 mainly consists of medicine names, entities in B2 obviously relate to chemical terms. 
\textbf{Last}, \PE can also put the ``nick name'' of a entity into the correct corresponding cluster. For example, \textit{Windy City} (i.e., Chicago) and \textit{Pittsburgh of the South} (i.e, Birmingham) were successfully recognised as names for cities.

\section{Related Work}

\paragraph{KGE techniques.}~~In recent years, a growing body of studies has been conducted on the matter of training KGEs. Roughly speaking, these KGE methods fall into two categories: distance-based models and semantic matching models. 

The line of researches regarding distance-based models, which measures plausibility of tuples by calculating distance between entities with additive functions, was initialised the KGE technique proposed by~\citet{TransE}, namely, TransE. 
After that, a battery of follow-ups have been proposed, including example models like TransH~\citep{TransH}, TransR~\citep{TransR}, and TransD~\citep{TransD}. These algorithms have enhanced ability on modelling complex relations by means of projecting entities into different (more complex) spaces or hyper-planes. More recently, a number of studies attempt to further boost the quality of KGEs through a way of adding orthogonality constraints~\citep{RotatE, ote} for maintaining the relation embedding matrix being orthogonal, which is also the paradigm we follow in the present paper (see \cref{sec:method}).

In contrast, semantic matching models measure the plausibility of tuples by computing the similarities between entities with multiplicative functions. Such an similarity function could be realised using, for example, a bilinear function or a neural network. Typical models in this line includes DistMult~\citep{DistMult}, ComplEx~\citep{ComplEx}, ConvE~\citep{ConvE}, TuckER~\citep{TuckER}, and QuatE~\citep{QuatE}.

\paragraph{Accelerating KGE training.}~~All those KGE approaches share the same issue of their low speed in both training and inference phases (see \citet{rossi2020knowledge} for a controlled comparison of the efficiency across different methodologies). In response to this issue, some state-of-the-art KGE algorithms attempted to accelerate their inference speed either through making use of the high-speed of the convolutional neural networks~\citep{ConvE} or through reducing the scale of parameters of the model~\citep{QuatE, DistilE}.

As for the acceleration of model training, a number of attempts have been conducted in a mostly engineering way. These well-engineered systems adopt linear KGE methods to multi-thread versions in other to make full use of the hardware capacity~\citep{joulin2017fast, han2018openke}, which accelerates training time of, for example, TransE, from more than an hour to only a couple of minutes. Nonetheless, this line of work has two major issues: one is that training models faster in this way does not necessarily mean they also emit less, as process scheduling of a multi-thread system can be energy-consuming. The other is that they are all extensions of linear KGE models only (also noting that linear models are naturally much faster than other non-linear models) without any algorithmic contribution, which leading to the performance of the resulting models limited by the upper bound of linear models (e.g., recent state-of-the-art methods in Tab.~\ref{tab:main}, such as RotH, are nonlinear approaches).

\section{Conclusion}

In this paper, we proposed a novel KGE training framework, namely \PE, which is eco-friendly, time-efficient and can yield very competitive or even near-state-of-the-art performance. Extensive experiments show that our method is valuable especially considering its significant and substantial reduction on training time and carbon footprint.

\section*{Broader Impact}

We provided a efficient KGE training framework in this paper. The resulting KGEs, akin to all previous KGE models, might have been encoded with social biases, e.g., the gender bias~\citep{fisher2019measuring}. We suggest this problem should always be looked at critically. For whoever tend to build their applications grounding on our KGEs, taking care of any consequences caused by the gender bias is vital since, in light of the discussion in~\citet{larson-2017-gender}, mis-gendering individuals/entities is harmful to users~\cite{10.1145/3274357}. 
Additionally, as having been proven in this paper, our method emits less greenhouse gases and therefore, has less negative environmental repercussions than any other KGE approaches.

\section*{Acknowledgements}
This work is supported by the award made by the UK Engineering and Physical Sciences Research Council (Grant number: EP/P011829/1) and Baidu,~Inc.
We would also like to express our sincerest gratitude to Chen Li, Ruizhe Li, Xiao Li, Shun Wang, and the anonymous reviewers for their insightful and helpful comments.

\bibliography{custom}
\bibliographystyle{acl_natbib}

\clearpage
\appendix

\renewcommand\thefigure{\thesection.\arabic{figure}}
\setcounter{figure}{0}
\renewcommand\thetable{\thesection.\arabic{table}}
\setcounter{table}{0}
\renewcommand\theequation{\thesection.\arabic{equation}}
\setcounter{equation}{0}

\section{Gram-Schmidt Process of \citet{ote}}\label{app:Gram-Schmidt}
The Gram-Schmidt process takes a set of tensor $S=\{v_1,\cdots,v_k\}$ for $k\leq {d_s}$ and generates an orthogonal set
$S'=\{u_1,\cdots,u_k\}$ that spans the same $k-$dimensional subspace of $\mathcal{R}^{d_s}$ as $S$, such that 
\begin{equation}
    u_i = \frac{t_i}{||t_i||}\,\, \mathrm {s.t.} \,\, t_i = v_k - \sum_{j=1}^{k-1}\frac{\langle v_k,t_j\rangle}{\langle t_j,t_j\rangle}t_j,
\end{equation}
where $t_1=v_1$ and $\langle v,t\rangle$ denotes the inner product of $v$ and $t$. Its complexity is $\mathcal{O}({d_s}^3)$ and the parallelisation is not trivial.

\section{Bandwidth Comparison}\label{app:bandwidth}
\begin{figure}[h]
  \centering
  \includegraphics[width=\columnwidth, trim={.4cm .4cm .2cm .3cm}, clip]{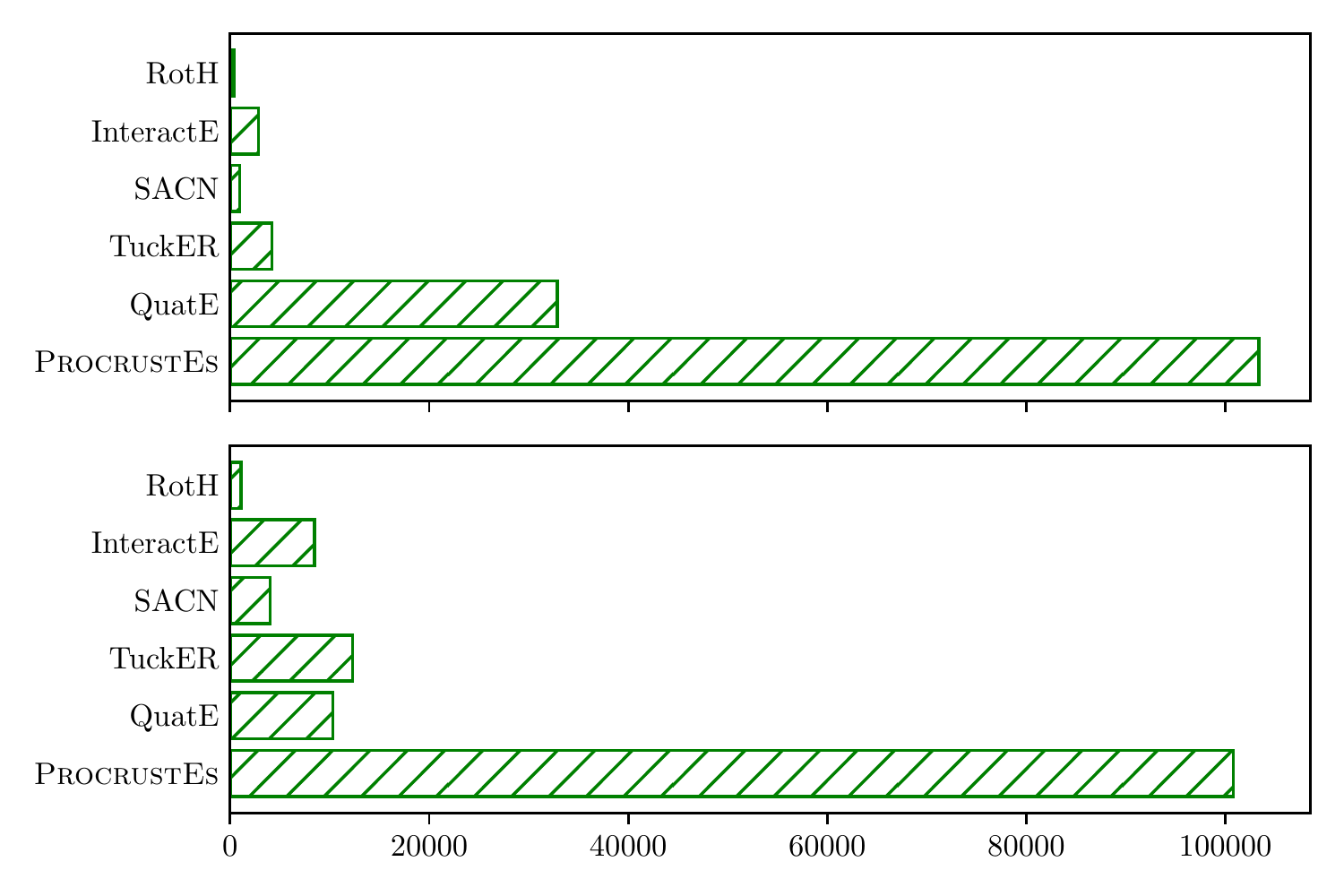}  
\caption{Comparison of bandwidth (number of processed samples per second). The upper and the lower are respectively for WN18RR and FB15k-237.}
\label{fig:bandwidth}
\end{figure}

To further ascertain the efficiency advantage of \PE by ruling out factors such as numbers of all epochs, we pick four frameworks with strongest MRR performance and estimate their bandwidth during training, as illustrated in Fig.~\ref{fig:bandwidth}.

We can see that although some baselines have been engineered for enhanced computational efficiency, e.g., by default RotH creates 24 threads for multiprocessing, on both datasets they still substantially underperform \PE  with huge margins in terms of bandwidth.

\pagebreak
\section{Visualisation Comparisons}\label{app:vis_w2v}

\begin{figure}[h]
\begin{subfigure}{\columnwidth}
  \centering
  \includegraphics[width=\columnwidth, trim={3cm 1.5cm 1cm 2.7cm}, clip]{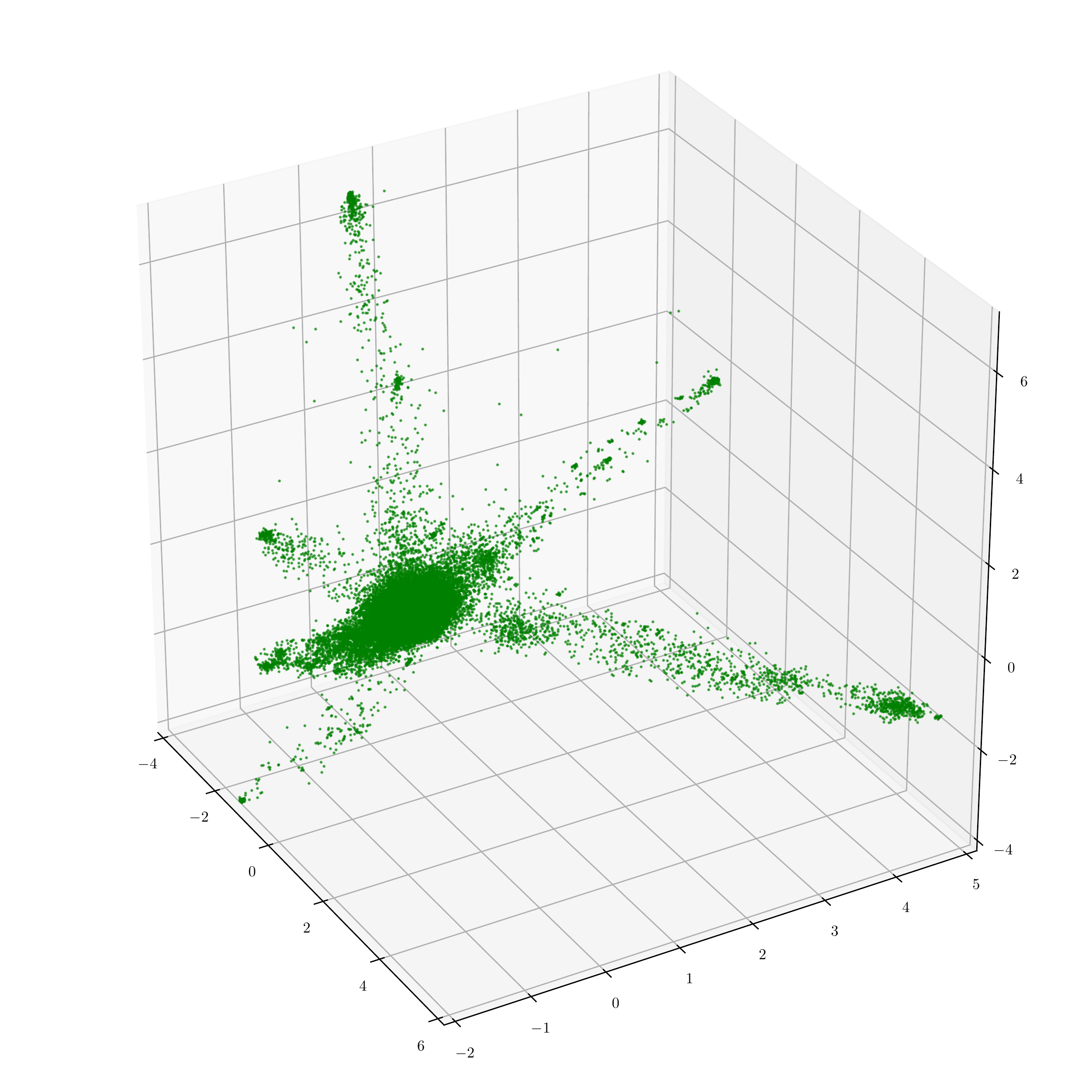}  
\caption{\PE entity embeddings (Fig.~\ref{fig:pca} with 3D axes and grids enabled)}\label{fig:3d-kge}
\end{subfigure}
\begin{subfigure}{\columnwidth}
  \centering
  \includegraphics[width=\columnwidth, trim={3cm 1.5cm 1cm 2.7cm}, clip]{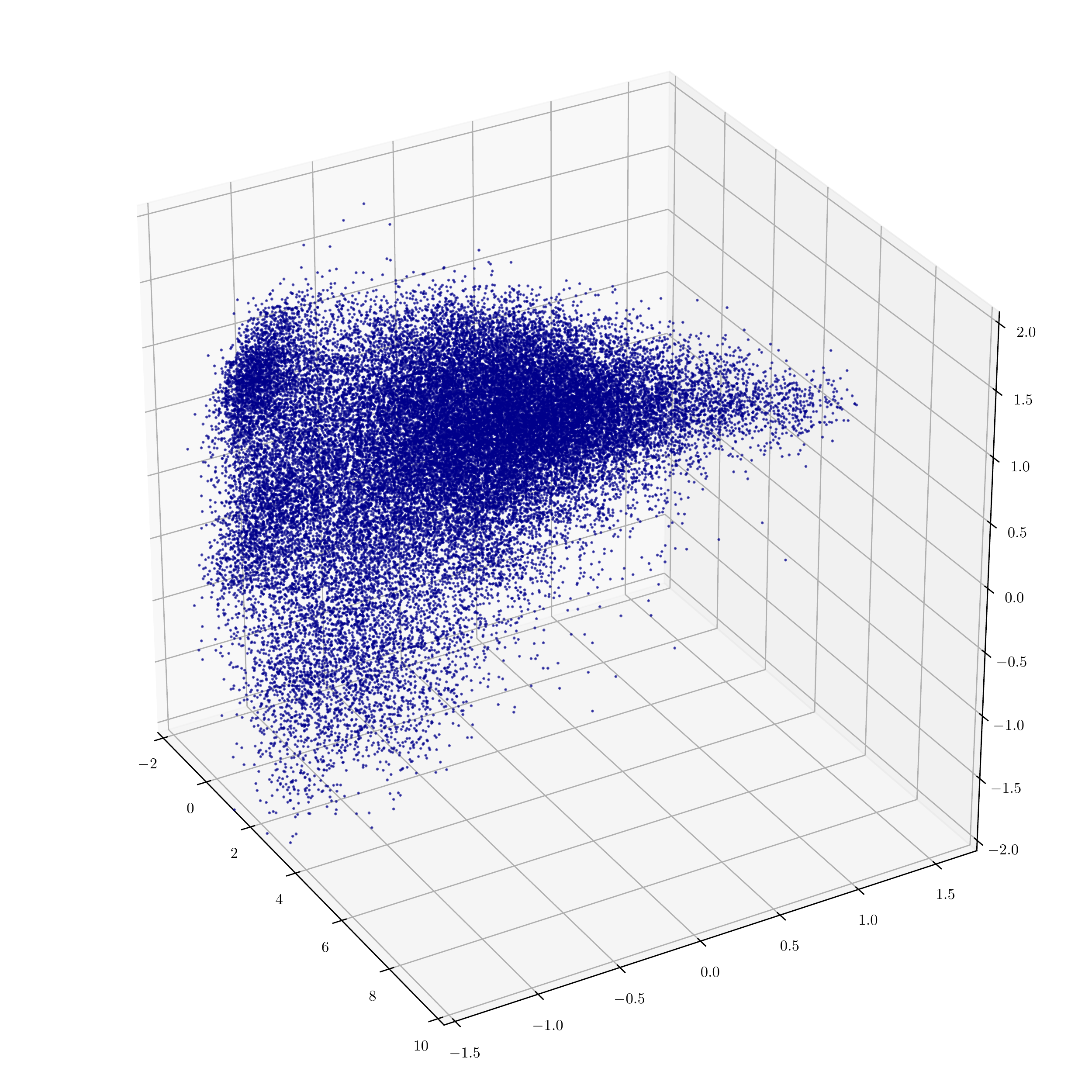}  
\caption{Word embeddings}\label{fig:3d-w2v}
\end{subfigure}
\caption{3D PCA visualisation of \PE entity embeddings and pretrained word embeddings (top 40K vectors of English Wikipedia Embeddings at \url{https://tinyurl.com/ft-wiki-vec}). It is obvious that they have very distinct typologies.}
\label{fig:3dvis}
\end{figure}

\end{document}